\pdfoutput=1  %
\documentclass[red]{preprint}

\usepackage{amsmath}
\usepackage{amssymb}
\usepackage{mathtools}
\usepackage{xspace}
\usepackage{xurl}      %
\usepackage{booktabs}
\usepackage{colortbl}
\usepackage{enumitem}
\usepackage{array}
\usepackage{multirow}
\usepackage{listings}
\usepackage{needspace}
\usepackage{wrapfig}
\usepackage{pifont}
\usepackage{marvosym}
\usepackage[capitalize,noabbrev,nameinlink]{cleveref}
\crefname{appendix}{Appendix}{Appendices}
\Crefname{appendix}{Appendix}{Appendices}

\setlist{nosep,leftmargin=*}

\newif\ifanon
\anonfalse
\newcommand{\anon}[2]{\ifanon #1\else #2\fi}

\newcommand{\boardurl}{\anon{[anonymized leaderboard URL]}{\url{https://huggingface.co/spaces/Diaugeia/TSFLab}}}

\newcommand{\method}{\textsc{TSFLab}\xspace}
\newcommand{\nmodels}{323\xspace}      %
\newcommand{\ndatasets}{93\xspace}     %
\newcommand{\nskills}{17\xspace}       %
\newcommand{\ncomponents}{71\xspace}   %
\newcommand{\ncomposed}{263\xspace}    %

\makeatletter
\DeclareRobustCommand\onedot{\futurelet\@let@token\@onedot}
\def\@onedot{\ifx\@let@token.\else.\null\fi\xspace}
\def\eg{e.g\onedot} 
 
\makeatother

\definecolor{codebg}{HTML}{F5F4EF}
\definecolor{codeframe}{HTML}{DDDBD2}
\definecolor{codeink}{HTML}{35558C}
\definecolor{codestring}{HTML}{0E7A55}
\definecolor{codecomment}{HTML}{8A8880}
\title{\raisebox{-0.15em}{\includegraphics[height=0.9em]{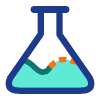}}\,TSFL{\fontsize{0.8em}{1em}\selectfont AB}: A Self-Improving
Agent-Agnostic Workbench for Time Series Forecasting Research}

\author{ChengAo Shen\textsuperscript{1}, Yanjun Zhao\textsuperscript{2}, Haoxin Liu\textsuperscript{3},
Bonnie Chen\textsuperscript{4}, Miles Ma\textsuperscript{5},\\
Dongjin Song\textsuperscript{6}, Hanghang Tong\textsuperscript{2}, Qingsong Wen\textsuperscript{7},
Jingrui He\textsuperscript{2}, Jingchao Ni\textsuperscript{1\Letter}}

\newcommand{\instlogo}[2]{%
  \sbox0{#2}%
  \raisebox{2.1mm}{\sffamily\bfseries\scriptsize\color{brandfg}#1}\kern1mm%
  \raisebox{-0.5\ht0}{\usebox0}}
\institutions{%
  \centering
  \instlogo{1}{\includegraphics[height=7mm]{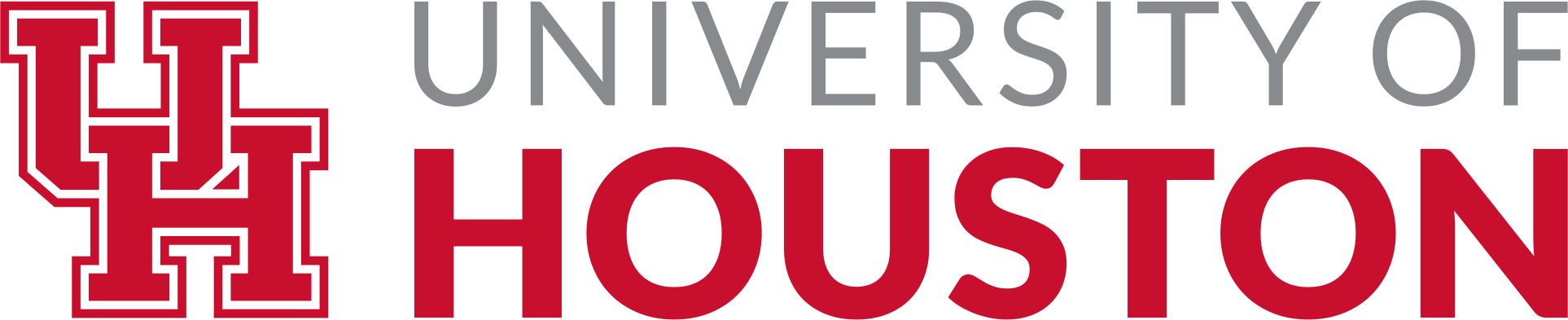}}\hspace{6mm}%
  \instlogo{2}{\includegraphics[height=7mm]{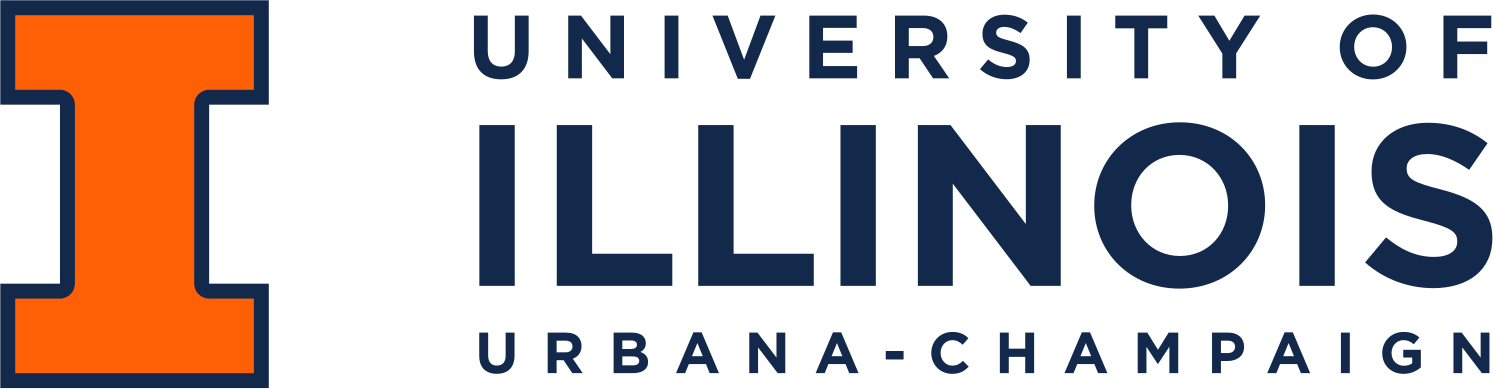}}\hspace{6mm}%
  \instlogo{3}{\includegraphics[height=6.5mm]{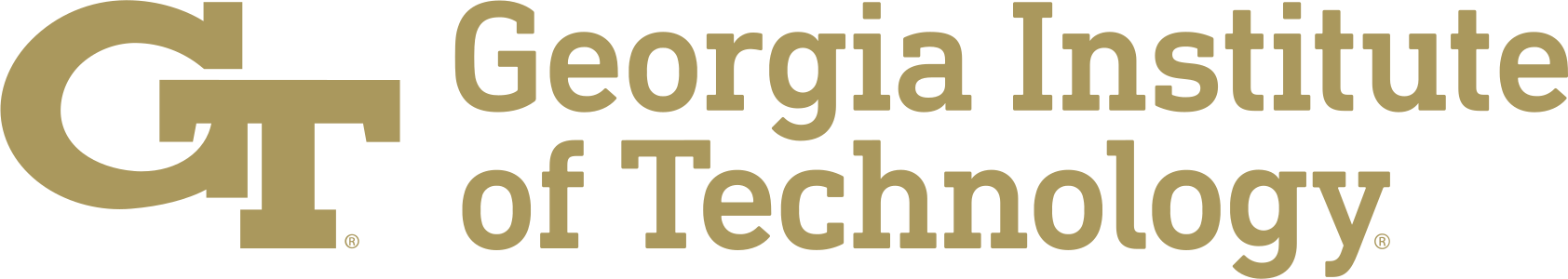}}\hspace{6mm}%
  \instlogo{4}{\includegraphics[height=8.5mm]{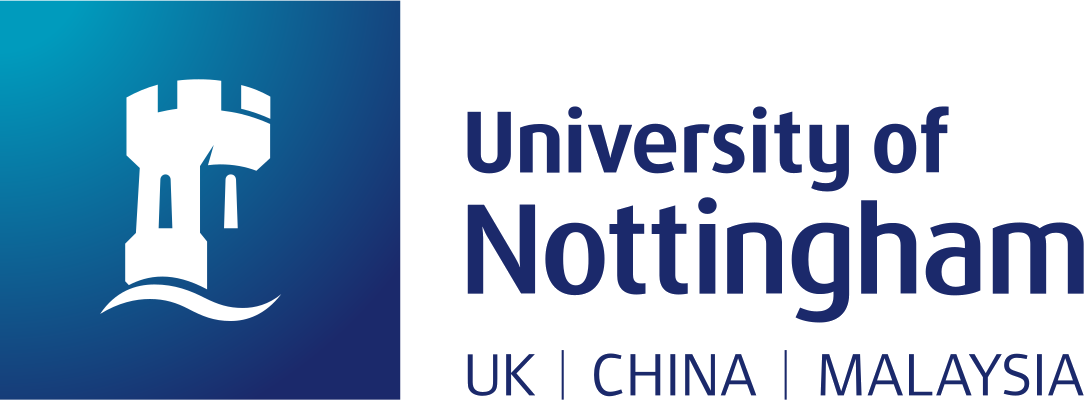}}\par\vskip 0.45cm  %
  \instlogo{5}{\includegraphics[height=5.5mm]{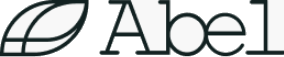}}\hspace{6mm}%
  \instlogo{6}{\includegraphics[height=6.5mm]{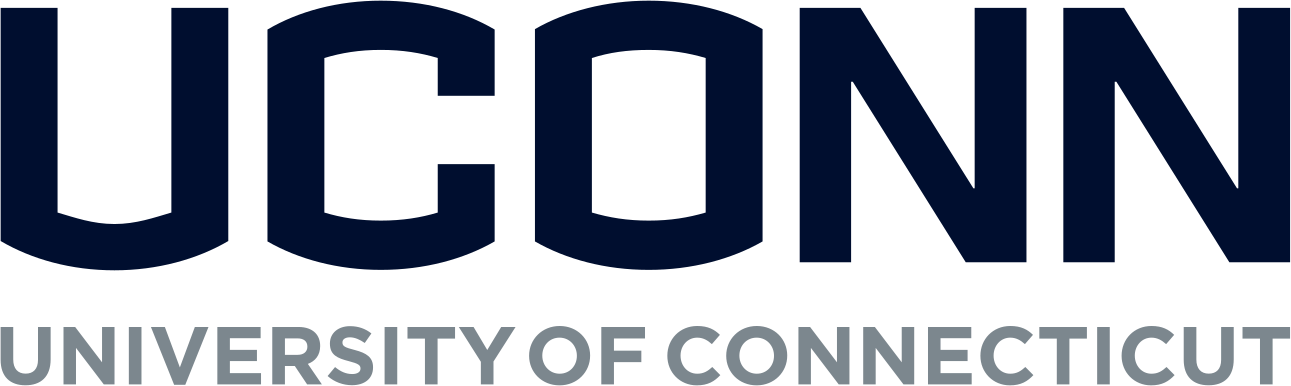}}\hspace{6mm}%
  \instlogo{7}{\includegraphics[height=7mm]{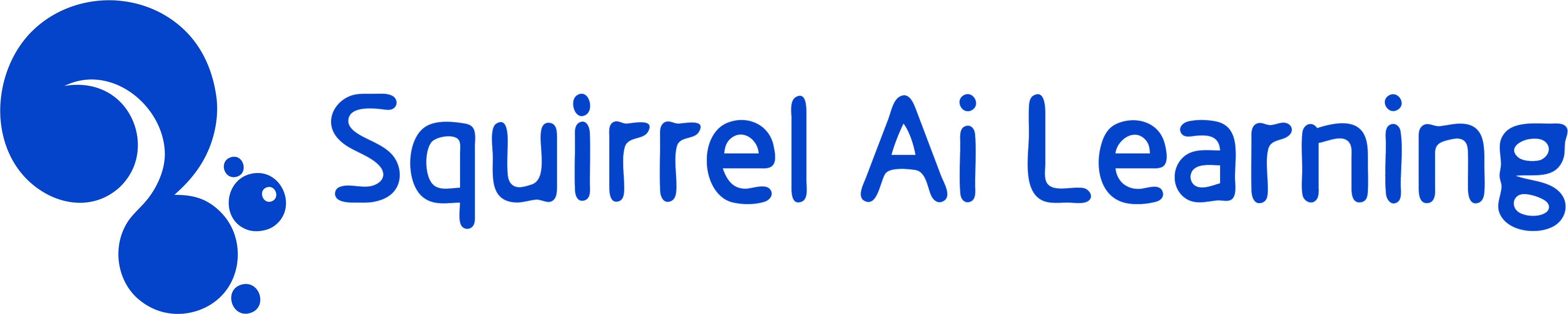}}\par}  %

\abstract{%
As publications and products on time series forecasting (TSF) proliferate,
researchers face a growing burden of analyzing existing work.
First, not all publications release code. For instance, only about 20\% of arXiv TSF
papers have a code link in the surveyed period. Second, released code also requires
manual maintenance.  Thus, TSF researchers must constantly adapt or implement existing works.
However, independent implementations can differ in experimental settings
and are difficult to maintain. We therefore
propose to use AI agents for benchmarking TSF
research and even supporting autoresearch, e.g., developing new TSF models.
Yet directly applying existing coding agents, such as
\texttt{Claude Code} and \texttt{Codex}, is suboptimal:
prohibitive costs, limited scalability to large research projects
(e.g., benchmarking), and lack of self-maintenance.
To address these challenges, we propose \method, a self-improving
agent-agnostic ``workbench'' that equips compatible coding
agents with shared resources and tools
for low-cost
TSF research and continuous maintenance
of
research resources.
Our experiments cover reproduction, benchmarking, and
new method development.
In reproduction and benchmarking, \method improves task completion by
20--163\% and cuts fresh tokens per completed task by 11--66\% across baselines.
A pilot study of agent-driven method development also shows improved
forecasting accuracy via \method, with 32.8--73.6\% fewer fresh tokens and
9.2--68.8\% less development time than the other settings.
}

\metadata[Code]{\url{https://github.com/D2I-Group/TSFLab}}

\begin{document}

\maketitle

\section{Introduction}
\label{sec:intro}

\begin{figure*}[t]
    \centering
    \includegraphics[width=\textwidth]{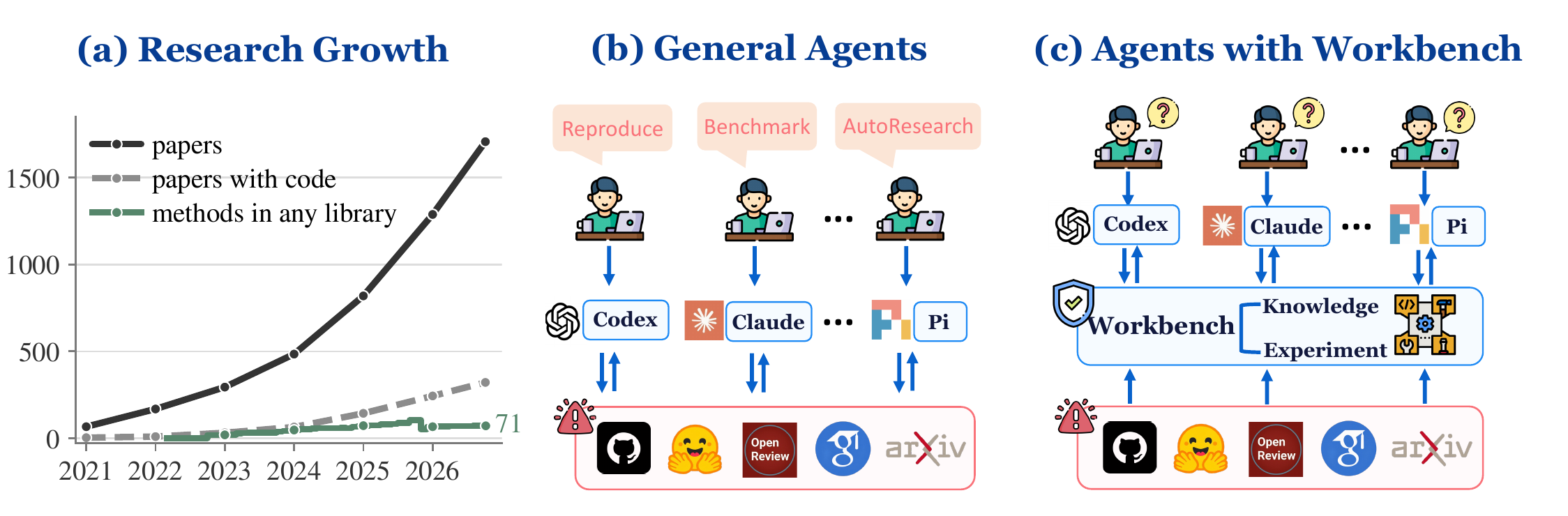}
    \caption{(a) Growth of TSF papers, papers with code, and methods in
        libraries. (b) General agents access external
        resources directly for each task. (c) Agents with a workbench share verified
        resources and reusable experiments.}
    \label{fig:intro}
    \vspace{-1em}
\end{figure*}

Time series forecasting (TSF) is a fundamental task in energy \citep{hong2016load},
transportation \citep{li2018dcrnn}, finance \citep{sezer2020financial},
weather \citep{lam2023graphcast}, and public health
\citep{cramer2022covid}. In Fig.~\ref{fig:intro}(a), we summarize the statistics of
TSF preprints and code releases using data from arXiv. As can be seen, code
releases grow much more slowly than papers: only about one in five of the
1,707 papers since 2020 provides a code link. Researchers,
particularly those who are new to this area, must often reimplement
baselines and build evaluation pipelines manually.
While TSF libraries exist, keeping them up to date is challenging:
\texttt{TSLib} \citep{tslib2026notice} grew from 18 to 41 methods despite over 1,200
new papers since 2024, and announced in April 2026 that it is no longer
maintained due to limited maintainer capacity.

AI agents offer a promising solution.
Agent-based autoresearch has shown potential in reproducing experiments
\citep{starace2025paperbench}, conducting new experiments
\citep{huang2024mlagentbench,nathani2025mlgym}, and developing new methods
\citep{lu2026aiscientist,schmidgall2025agentlab}.
In domains such as biology and chemistry, systems such as
\texttt{ChemCrow} \citep{bran2023chemcrow} and
\texttt{PantheonOS} \citep{xu2026pantheonos} equip agents with domain-specific
knowledge, tools, and workflows, accelerating research in these areas.
In these agents, the infrastructure surrounding their LLMs is called a
``harness'' \citep{anthropic2025harness,lee2026metaharness}.

Unlike a harness, which is part of each agent, a ``workbench'' lies outside
agents: it is reusable and shareable across agents. Workbenches exist for
general science, such as \texttt{ToolUniverse} \citep{tooluniverse2025} and
\texttt{Scientific Agent Skills} \citep{kassis2026scientificskills}, and
for biomedicine, such as \texttt{Biomni-E1} \citep{huang2026biomni}, but not
for TSF. We introduce a domain-specific workbench that supports AI agents
throughout TSF research.
Existing TSF agent research focuses on harness development
\citep{zhao2025tsci,timecopilot2025,shen2026metacaster}.
However, once their systems are developed, they cannot automatically benefit
from the emergence of newer agents.
In contrast, a workbench is plug-and-play infrastructure for any compatible agent, such
as \texttt{Claude Code} \citep{anthropic2026claudecode}, \texttt{Codex} \citep{openai2026codex},
and \texttt{Pi} \citep{pi2026agent}, benefiting from their continuously
improving capabilities. As compared in Fig.~\ref{fig:intro}(b) and (c), a workbench
enables reuse across agents and teams: a reproduction agent can add verified implementations to the workbench for
a benchmarking agent to reuse. Another agent can then reuse these benchmarks for method development.

Using agents with only their harness for TSF research presents two challenges:
(1)~\textbf{Slowly Updated TSF Knowledge.} An LLM-based agent's knowledge depends
on its training cutoff, so it may not cover newer TSF models and experimental
protocols \citep{qiu2024tfb,brigato2026champions}. Web search alleviates this
problem, but repeated searches for large-scale research projects are
token-costly. Retrieved sources may be incomplete or unreliable, and unchecked
errors may propagate into subsequent research.
For example, the official code of \texttt{Informer} (i.e., a TSF model)
\citep{zhou2021informer} mistakenly dropped the last test batch in evaluation,
which has influenced the settings of many subsequent studies \citep{qiu2024tfb}.
(2)~\textbf{Inconsistent Research Settings.} As shown in Fig.~\ref{fig:intro}(b),
agents that configure experiments independently may resort to different
information sources, leading to different settings, implementations, and
incomparable results. This makes it hard to share findings across agents and
teams, and leads to unnecessary repeated experiments
\citep{wang2024voyager,amac2026}.
Both challenges make agent-based TSF research costly and hard to scale:
agents repeat source searches, model implementations, and experiments
across agents and teams.

To address these challenges, we propose \textbf{\method}, a novel TSF workbench, as illustrated in Fig.~\ref{fig:intro}(c).
\method\ has four modules: (a)~\textbf{TSF-Domain Guide} enables TSF-awareness
through common research workflows, experimental settings, evaluation
protocols, and known pitfalls;
(b)~\textbf{Knowledge Structurer} decomposes existing TSF models into reusable
units and organizes these units, methods, and diverse benchmark datasets into a
unified structure. This structure is realized by a ``knowledge card'' for each
method, unit, and dataset, read level by level (summary, card, source) as a
decision requires, so that AI agents can reason efficiently without
token-costly reading of code and data details.
(c)~\textbf{Experiment Workflower} executes agent-declared experiments through
a unified runner that validates and runs experiment configurations and reports
results under consistent protocols.
(d)~\textbf{Recursive Improver} routinely checks and verifies up-to-date TSF
research outcomes and integrates verified artifacts (e.g., models,
datasets) into the workbench for self-maintenance. The workbench is updated
independently of agent releases.
Our contributions are as follows.
\begin{itemize}[noitemsep,topsep=0pt,leftmargin=*]
    \item We identify two causes of agent inefficiency in TSF: slowly updated knowledge and inconsistent settings.
    \item We propose \method, an agent-agnostic TSF workbench that empowers any compatible agent without altering it.
    \item Using \method, we integrated \nmodels TSF models and compiled
          \ndatasets datasets, along with \ncomponents model units that can be
          reused for autoresearch.
    \item We evaluate \method on three research tasks. In reproduction and
          benchmarking, it raises completion by 20--163\% and cuts fresh tokens per completed task
          by 11--66\% across baselines. A pilot study explores method development with
          fewer fresh tokens and less development time.
\end{itemize}

\section{Related Work}

\noindent{\textbf{TSF Libraries and Benchmarks}}.
\texttt{TSLib} \citep{wang2026tslib}, \texttt{BasicTS} \citep{shao2025basicts},
\texttt{TFB} \citep{qiu2024tfb}, and \texttt{NeuralForecast}
\citep{nixtla2023neural} provide implementations and evaluation
pipelines, and \texttt{GIFT-Eval} \citep{aksu2024gifteval} benchmarks time
series foundation models. Results and rankings change with evaluation
settings and tuning budgets \citep{qiu2024tfb,brigato2026champions}, yet
agents must read library code to recover these settings. Methods are added
and maintained by hand, so coverage lags the literature
(Fig.~\ref{fig:intro}(a)). \method describes its \nmodels methods and
\ndatasets datasets in knowledge cards read level by level, with more
methods than these libraries (Fig.~\ref{fig:further}).

\noindent{\textbf{Research Agents and Harnesses}}.
Agents show potential to reproduce papers \citep{starace2025paperbench}, run
machine learning experiments \citep{huang2024mlagentbench,nathani2025mlgym},
and propose new methods \citep{lu2026aiscientist,schmidgall2025agentlab},
but rely on general training knowledge.
Domain agents such as \texttt{ChemCrow} \citep{bran2023chemcrow} and
\texttt{PantheonOS} \citep{xu2026pantheonos} add chemistry and genomics
tools. TSF agents such as \texttt{TimeSeriesScientist} \citep{zhao2025tsci},
\texttt{TimeCopilot} \citep{timecopilot2025}, and \texttt{MetaCaster}
\citep{shen2026metacaster} orchestrate forecasters or optimize their training: they produce forecasts rather than conduct TSF
research. In these systems, domain support lives in the agent's
harness \citep{anthropic2025harness,lee2026metaharness}, so it serves one
agent and must be rebuilt for newer ones. \method serves any compatible agent
unchanged.

\noindent{\textbf{Agent Workbenches}}.
Workbenches move domain support outside agents. \texttt{ToolUniverse}
\citep{tooluniverse2025} exposes scientific tools to different agents and
generates new ones, \texttt{Scientific Agent Skills}
\citep{kassis2026scientificskills} packages portable procedures, and
\texttt{Biomni-E1} \citep{huang2026biomni} supplies biomedical tools,
software, and databases. They provide tools and procedures for general science
or biomedicine, rather than TSF methods under declared evaluation protocols. \method is a TSF
workbench with TSF-Domain Guide, Knowledge Structurer, and
Experiment Workflower, which executes comparisons under shared, explicit
evaluation protocols across agents and teams.

\noindent{\textbf{Self-Improving Agents}}.
\texttt{STOP} and the \texttt{Darwin G\"odel Machine} revise agent code through
score-based selection \citep{zelikman2024stop,zhang2025dgm}, \texttt{Voyager}
accumulates skills \citep{wang2024voyager}, and memory systems retain
information across tasks \citep{amac2026,consistencygate2026}. These methods
improve a single agent. The Recursive Improver of \method improves the shared
workbench instead: it checks each method against its paper and official code
and merges verified methods, units, datasets, and findings, so research
outputs from every task feed both the improvement and the maintenance of
the workbench.

\section{\method}

\begin{figure*}[t]
    \centering
    \includegraphics[width=\textwidth]{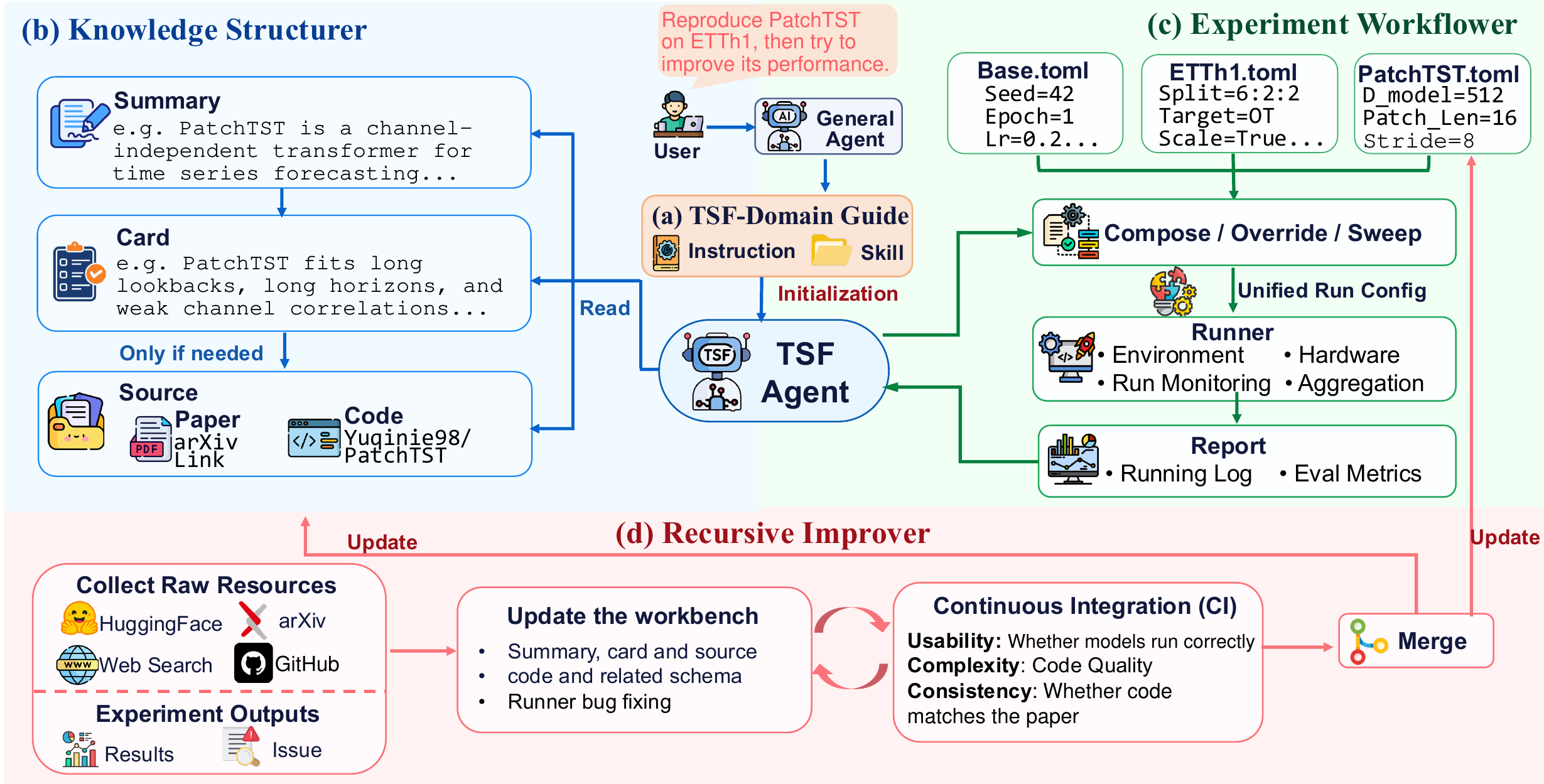}
    \caption{Overview of how an agent works with the \method workbench.}
    \label{fig:overview}
\end{figure*}

\subsection{Problem Statement}

Let $\mathcal{W}$ be a workbench, %
and $\Pi$ be the set of
agents that are compatible with
$\mathcal{W}$.
Each $\pi\in\Pi$ %
comprises its underlying model and harness.
The harness coordinates model and tool execution, manages context, and
organizes internal workflows.
In contrast, the workbench $\mathcal{W}$ provides
reusable domain knowledge, research procedures, and experimental tools
without changing the model or harness:
\begin{equation}
    \pi^*=\operatorname{Augment}(\pi,\mathcal{W}),\qquad \pi\in\Pi.
\end{equation}
Our goal is to improve valid task completion (i.e., completing a research task exactly as required) and reduce cost per completed
task. We describe the evaluation metrics
in \S\ref{sec:setup}.

\method supports research through four modules, as shown in Fig.~\ref{fig:overview}:
(a)~\textbf{TSF-Domain Guide} is used by an agent for task procedures,
(b)~\textbf{Knowledge Structurer}
selects methods,
configures parameters,
and
reuses model units,
(c)~\textbf{Experiment Workflower}
executes the declared
experiments, and (d)~\textbf{Recursive Improver} verifies new resources and outcomes
for
updating the shared workbench.

\subsection{TSF-Domain Guide}

TSF research requires domain knowledge of data integration, method
implementation, and evaluation protocols (\S\ref{sec:intro}). TSF-Domain
Guide connects this knowledge to workbench procedures: {\em entry instructions}
introduce conventions and module interactions, while \nskills task-specific
skills provide steps, checks, and pitfalls as needed.

For example, when benchmarking methods on
a new dataset, a skill guides the
agent to identify time fields, forecasting variables, sampling frequency,
and evaluation settings, then register data through the workbench's ``add''
interface and inspect partitioning and scaling through its ``split'' interface.
The agent
does not need to maintain its own loading and splitting code. The skill
also requires checking whether forecasts are evaluated on one target or
all variables, keeping metrics consistent across methods. These choices
establish the constraints for resource selection and configuration. The entry instructions and task-specific skills are detailed in
App.~\ref{app:guidance-source}.

\subsection{Knowledge Structurer}
\label{sec:structure}

\begin{wraptable}[12]{r}{0.5\textwidth}
    \vspace{-1.2em}
    \centering
    \small
    \setlength{\tabcolsep}{3pt}
    \renewcommand{\arraystretch}{1.12}
    \begin{tabular}{@{}>{\centering\arraybackslash}m{0.17\linewidth}>{\raggedright\arraybackslash}m{0.35\linewidth}>{\raggedright\arraybackslash}m{0.42\linewidth}@{}}
        \toprule
        \textbf{Level} & \textbf{When to consult}                                            & \textbf{Information provided}                                              \\
        \midrule
        Summary        & Screening candidates                                                & Category and use cases                                                      \\
        \midrule
        Card           & Resource selection, configuration, unit reuse & Capabilities and limits, parameters and interfaces, verification records \\
        \midrule
        Source         & Checking details, resolving doubts, modifying methods & Original papers, reference code, detailed evidence \\
        \bottomrule
    \end{tabular}
    \caption{Three levels of research knowledge.}
    \vspace{-1em}
    \label{tab:knowledge-levels}
\end{wraptable}

After establishing task requirements, the agent uses Knowledge Structurer
to select and configure resources. It provides a knowledge card for each method, dataset, and unit, organized into summary, card, and source levels.
Summaries support screening; cards describe capabilities, parameters,
interfaces, and limitations; sources provide papers, reference code, and
detailed explanations. This progressive disclosure
\citep{pd2026longcontext,pd2026skills} lets agents retrieve the detail needed
for each decision without rereading sources
(Tab.~\ref{tab:knowledge-levels}).

Method cards connect mechanisms to use cases and configuration. For
example, the card of \texttt{PatchTST} \citep{nie2023patchtst} explains its
channel-independent design---modeling variables separately with shared
weights---so an agent can compare it with methods that learn cross-variable
interactions. It also states whether the implementation supports exogenous
inputs or probabilistic forecasts, clarifying task compatibility. Parameter descriptions define patch length and stride; existing settings
and input-window constraints help agents exclude unsuitable configurations
and plan validation comparisons.
The workbench sets data-dependent parameters, such as input channel count,
from declared dependencies. Dataset cards record data characteristics
and split, scaling, and evaluation protocols. Thus, agents screen resources
by both forecasting mechanism and task compatibility, consulting source
details when a configuration or implementation decision requires them.

Knowledge Structurer also extracts common blocks, such as
\texttt{RevIN}, series decomposition, and attention variants, into shared
units that methods import, enabling shared maintenance; \ncomposed methods
reuse at least one unit. Each unit has a card, and method cards record
which units they use. For example, \texttt{RevIN} removes an input window's level
and scale and restores them in the forecast. Its card describes the shifts
it addresses, parameters, and channel requirements, helping agents decide
whether to reuse it in a new method.

\subsection{Experiment Workflower}
\label{sec:experiment}

With resources selected, Experiment Workflower executes agent-declared
experiments through a unified
runner. Each method's specification defines parameter requirements,
construction, and input and output interfaces, allowing the runner to
instantiate the model and connect it to shared loading, training, and
evaluation routines. Presets supply training settings, dataset protocols, and architecture
parameters for reuse across experiments.

Agents use \textbf{Compose} to combine presets, \textbf{Override} to change settings, and \textbf{Sweep} to expand methods or values.
For new-data benchmarking, an agent combines dataset and method presets,
fixes the lookback, horizon, and training budget, and sweeps methods and
seeds. The runner expands configurations into a run matrix, then checks
configuration validity, data, model compatibility, environments, and budgets.
It returns problems before training, connecting the agent's declaration to shared training and evaluation.

Shared routines split data as configured, fit scaling on training data only,
select checkpoints on validation data, and compute metrics consistently.
The runner schedules runs within resources and budgets, preserving status,
checkpoints, and prior attempts. Recovery checks configuration, data, and
code fingerprints, skips completed runs, and resumes failed runs from their
last complete epoch. Status, wait, cancel, and resume interfaces reduce
manual management of scripts, processes, and logs.

The runner returns metrics, training time, peak GPU memory, status, and
artifact paths, archiving configurations, seeds, environments, and data
versions. Result summaries compare methods only under the same data
processing, training, and evaluation protocol, reporting seed coverage and
variation. Recorded evidence supports subsequent comparisons, development, and
review through Recursive Improver (\S\ref{sec:improve}).

\subsection{Recursive Improver}
\label{sec:improve}

Here, self-improvement means incorporating verified research outputs and
fixes into the workbench's shared resources and workflows so later agents can reuse them. For example, evaluation problems found during
research sessions lead to revised guidance and runner checks; later sessions inherit these repairs.

Recursive Improver brings external resources and agent-produced outcomes
back into the shared workbench through checks
for usability, complexity, and consistency (Fig.~\ref{fig:overview}d).
Failed checks prompt revision
and rechecking before integration, supporting workbench maintenance.

\begin{table*}[t]
    \centering
    \begingroup
    \small
    \renewcommand{\arraystretch}{1.12}
    \setlength{\tabcolsep}{3.2pt}
    \begin{tabular*}{0.95\textwidth}{@{\extracolsep{\fill}}l r rrr rrr@{}}
\toprule
\textbf{Setting} & Completed$\uparrow$ & Time (h)$\downarrow$ & Tokens (M)$\downarrow$ & Calls$\downarrow$ & $\Delta$LOC$\downarrow$ & $\Delta$Funcs$\downarrow$ & $\Delta$CC$\downarrow$ \\
\midrule
\rowcolor{blue!6}
\multicolumn{8}{l}{\textit{Paper reproduction}} \\
\method & \textbf{12/12} & \textbf{1.96} & \textbf{0.434} & \underline{159.67} & \underline{148.08} & \textbf{8.25} & \textbf{25.83} \\
TSLib & \underline{10/12} & 2.61 & 0.533 & 164.60 & 160.67 & \underline{9.67} & \underline{31.33} \\
TFB & 7/12 & \underline{2.46} & \underline{0.487} & \textbf{131.86} & 153.42 & 11.58 & 36.00 \\
No workbench & 8/12 & 3.50 & 0.706 & 212.75 & \textbf{131.00} & 11.00 & 34.00 \\
\midrule
\rowcolor{green!6}
\multicolumn{8}{l}{\textit{Benchmarking on new data}} \\
\method & \textbf{42/48} & \textbf{0.05} & \textbf{0.022} & \textbf{9.05} & \textbf{0.00} & \textbf{0.00} & \textbf{0.00} \\
TSLib & \underline{30/48} & \underline{0.06} & \underline{0.035} & \underline{12.50} & 180.33 & \underline{9.17} & 52.17 \\
TFB & 16/48 & 0.08 & 0.065 & 18.88 & 345.67 & 30.33 & 76.50 \\
No workbench & 26/48 & 0.08 & 0.051 & 15.85 & \underline{144.00} & 16.33 & \underline{50.17} \\
\bottomrule
\end{tabular*}

    \par\endgroup
    \caption{Main results using \texttt{Codex CLI} with GPT-6 Luna. Costs are normalized by completed outcomes. Best results are bold; second-best are underlined. Metric definitions are in App.~\ref{app:measure}.}
    \label{tab:main}

\end{table*}

\noindent\textbf{Usability.}
Continuous integration (CI) checks basic operations. Method tests
cover construction, a training step, input and output requirements, and
saving and loading; dataset checks cover schema and evaluation protocols;
experiment records require complete configurations and runner-produced
results. They also confirm required card fields and agreement between
method specifications and presets, exposing integration errors.

\noindent\textbf{Complexity.}
CI checks function-level cyclomatic complexity and method size. Cyclomatic complexity counts independent execution paths through a
function; method size counts lines of code. These flag implementations that are harder to understand, test, and maintain.
Code above the limits requires simpler logic, smaller functions,
or shared units, then renewed runtime and behavioral checks.

\noindent\textbf{Consistency.}
An independent read-only review agent checks whether cards
describe implementations and whether implementations preserve the defining
mechanisms in papers and pinned official code, using papers alone when code
is unavailable. For example, it checks for cross-variable computation in
a method described as channel-independent. Paper--code setting differences and implementation choices are recorded.
For extracted units, automated tests compare parameters, outputs, and
input gradients before and after extraction, checking model state, forward computation, and backpropagation behavior on reference cases.

Accepted resources and verification records enter the workbench to support
subsequent benchmarking, reproduction, and method development. Additions
and fixes from these tasks undergo the same checks, forming a shared update
cycle that reduces repeated collection, implementation, and setup. Consequently, a session can contribute a reusable
method, unit, or workflow repair alongside its experimental results.

\section{Experiments}

\renewcommand{\dbltopfraction}{0.9}
\setcounter{dbltopnumber}{3}

\begin{figure*}[t]
\centering
\includegraphics{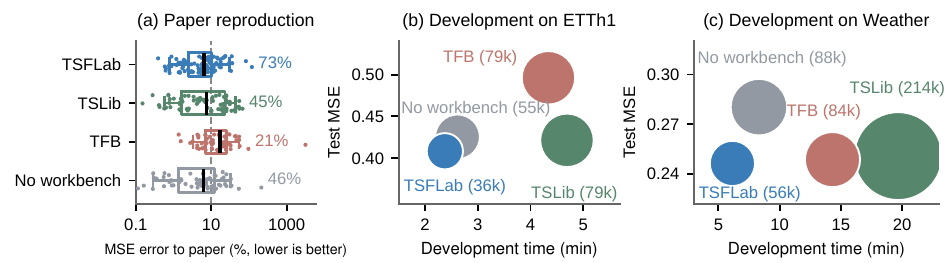}
\caption{Research-task quality. (a) Reproduction MSE error (log scale; lower is better); labels show shares within 10\% (96 scheduled). (b,c) Development test MSE versus time; bubble areas and labels show fresh tokens.}
\label{fig:compare}
\end{figure*}

We report completion, cost, and code changes (\S\ref{sec:main}), ablations (\S\ref{sec:ablation}), and coverage and maintenance (\S\ref{sec:further}).

\subsection{Experimental Setup}
\label{sec:setup}

\noindent\textbf{Baselines.}
We compare \method with two TSF codebases and with no workbench:
Time-Series-Library (\texttt{TSLib}) \citep{wang2026tslib}, a model library, and \texttt{TFB}
\citep{qiu2024tfb}, a benchmarking framework. Without a workbench, the agent
starts from an empty directory with PyTorch. All main settings use \texttt{Codex CLI} with GPT-6 Luna and identical
prompts, data, network access, and budgets; Tab.~\ref{tab:agents} varies the agent.

\noindent\textbf{Research Tasks.}\label{sec:tasks}
(i)~Paper reproduction: the agent reproduces the reported results of a recent
forecasting paper that has no official code and is absent from all compared
codebases (RhyMix \citep{shevtekar2026rhymix},
StateFlow \citep{gharwi2026stateflow}, AdaRDiff \citep{laglil2026adardiff},
and CoRe \citep{lin2026core}; App.~\ref{app:tasks}).
(ii)~Benchmarking on new data: given two new datasets, ERCOT and Openmeteo,
the agent benchmarks eight methods under a common protocol: three established
baselines (\texttt{DLinear} \citep{zeng2023transformers}, \texttt{PatchTST}
\citep{nie2023patchtst}, and \texttt{iTransformer} \citep{liu2024itransformer}),
two alignment and retrieval methods (TimeAlign \citep{hu2026timealign} and
GTR \citep{cao2026gtr}), and three methods for temporal heterogeneity and
embedding enhancement (TimeMosaic \citep{ding2026timemosaic}, PHAT
\citep{ma2026phat}, and EMAformer \citep{zhang2026emaformer}).
(iii)~Method development pilot: on ETTh1 and Weather, the agent searches for a
forecasting method using validation data, then evaluates its final selection
once on test data (App.~\ref{app:tasks}).

\noindent\textbf{Evaluation.}
Tab.~\ref{tab:main} reports completion, cost, and added code. Completed
counts, for reproduction, sessions whose results pass review (four papers
$\times$ three attempts); for benchmarking, method results that are
faithfully implemented and evaluated under the protocol (eight methods
$\times$ two datasets $\times$ three attempts). Method development results are shown
in Fig.~\ref{fig:compare}b,c, using test MSE, session time, and fresh tokens.
Time, Tokens (fresh tokens, excluding cache reads), and Calls (agent tool
calls) sum all attempts, including failures, and divide by the completed
count. $\Delta$LOC, $\Delta$Funcs, and $\Delta$CC are the net code lines,
added functions, and added cyclomatic complexity per session. Details are in
App.~\ref{app:measure}.

\begin{table*}[t]
\centering
\begingroup
\small
\renewcommand{\arraystretch}{1.12}
\setlength{\tabcolsep}{3.2pt}
\begin{tabular*}{0.95\textwidth}{@{\extracolsep{\fill}}l r rrr rrr@{}}
\toprule
\textbf{Setting} & Completed$\uparrow$ & Time (h)$\downarrow$ & Tokens (M)$\downarrow$ & Calls$\downarrow$ & $\Delta$LOC$\downarrow$ & $\Delta$Funcs$\downarrow$ & $\Delta$CC$\downarrow$ \\
\midrule
Full \method & \textbf{12/12} & \textbf{1.96} & 0.434 & 159.67 & 148.08 & \textbf{8.25} & \underline{25.83} \\
\midrule
\rowcolor{blue!6}
\multicolumn{8}{l}{\textit{(a) Workbench Module Ablation}} \\
w/o TSF-Domain Guide & 3/12 & 5.73 & 1.129 & 348.33 & \underline{147.33} & 11.33 & 39.42 \\
w/o Knowledge Structurer & 8/12 & \underline{2.08} & 0.438 & 118.75 & 171.42 & 9.42 & 31.58 \\
w/o Experiment Workflower & 5/12 & 2.24 & 0.533 & 153.00 & 178.75 & 10.42 & 36.42 \\
\midrule
\rowcolor{green!6}
\multicolumn{8}{l}{\textit{(b) Agent Ablation}} \\
Claude Code, Sonnet 5.5 & \underline{11/12} & 2.38 & \textbf{0.195} & \textbf{78.00} & 212.17 & 13.50 & 36.25 \\
Codex CLI, GPT-6 Sol & \textbf{12/12} & 2.88 & \underline{0.394} & \underline{82.25} & \textbf{132.17} & \underline{9.25} & \textbf{23.67} \\
\bottomrule
\end{tabular*}

\par\endgroup
\caption{Module and agent ablations. The full workbench uses \texttt{Codex CLI} with GPT-6 Luna. Best values are bold; second-best distinct values are underlined, including ties.}
\label{tab:ablation}\label{tab:agents}
\end{table*}

\subsection{Main Results}
\label{sec:main}

Tab.~\ref{tab:main} and Fig.~\ref{fig:compare} compare \method with the other
settings on reviewed task completion, cost, and code changes. Reproduction
accuracy is evaluated separately by relative error to the reported MSE. We highlight five
observations.
(1) \method completes the most tasks with the most accurate results. It
completes all 12 reproduction attempts and 42 of 48 benchmark results, the
most of any setting. Fig.~\ref{fig:compare}a shows that its reproductions are
also the closest to the papers: \method reproduces 73\% of the 96 scheduled
results within 10\% of the paper's MSE, against 21--46\% for the other
settings. Completion and agreement with the reported MSE thus provide
complementary views of reproduction quality.
(2) \method also costs less: each completed outcome takes fewer tokens and
less time in reproduction, and fewer tokens and tool calls in benchmarking,
consistent with agents using shared operations instead of rebuilding them.
(3) Its added code is simpler. Benchmarking needs no code changes, since
all eight methods are already integrated and Experiment Workflower
runs them from configuration; in
reproduction, agents add a similar number of lines but 15--29\% fewer
functions and 18--28\% less added cyclomatic complexity than the other
settings, consistent with reuse of existing units.
(4) An existing codebase alone is not enough. \texttt{TFB} completes fewer
outcomes than no workbench in both tasks, mostly because agents replace
requested methods with substitutes. A framework helps an agent only when the
agent can extend it faithfully.
(5) The method-development pilot shows similar benefits. On ETTh1 and Weather
(Fig.~\ref{fig:compare}b,c), \method reaches the lowest test MSE with less
time and fewer fresh tokens than every other setting; details are in
App.~\ref{app:tasks}.
Reproduction benefits from shared units while still requiring new model
code; benchmarking reuses complete methods. The latter therefore measures
the combined benefit of catalog availability and shared execution.
The six unaccepted benchmark results show that integrated methods alone
do not guarantee valid evaluation. Zero added code measures implementation
reuse; completion review establishes whether experiments follow the
declared evaluation protocol.

\begin{figure*}[!t]
\centering
\includegraphics[width=\textwidth]{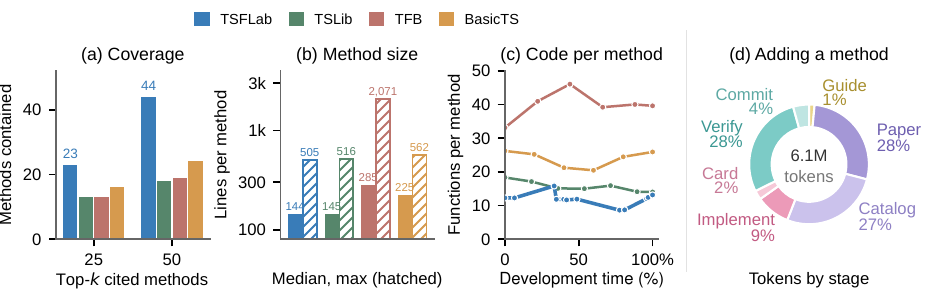}
\caption{Workbench coverage and evolution. (a) Top-cited coverage. (b) Median/max model lines. (c) Functions per method, initial (0\%) to latest (100\%). (d) Intake tokens including cache reads.}
\label{fig:further}
\end{figure*}

\subsection{Ablation Study}
\label{sec:ablation}

Tab.~\ref{tab:ablation} reports two groups of ablations. In (a), we remove one
module at a time on four reproduction tasks with three attempts each: the
TSF-Domain Guide (entry instructions and skills), the Knowledge Structurer
(cards and sources), or the Experiment Workflower (runner and experiment
commands). Recursive Improver is examined through maintenance and
refactoring outcomes in \S\ref{sec:further}. In (b), we replace the
agent: \texttt{Claude Code} with Sonnet 5.5, and \texttt{Codex CLI} with
GPT-6 Sol.

From Tab.~\ref{tab:ablation}(a), removing any module lowers completion, and
each fails differently. Without the guide, completion drops
the most: agents write their own data loaders, and every rejected attempt
uses wrong ETTh1 split boundaries, so the guide mainly keeps agents on the
evaluation protocol. Without the Knowledge Structurer, cost stays close to the
full workbench, but agents reuse fewer units and add more code and
complexity. Without the Experiment Workflower, agents write their own training
and evaluation code; they add the most code, and their errors move into
evaluation, such as tuning on test scores and averaging batch errors. Shared
execution thus removes errors that finite results alone do not reveal.
Similar token use without the Knowledge Structurer, despite lower completion, shows that token cost alone cannot measure reuse.

\label{sec:agents}
Tab.~\ref{tab:ablation}(b) highlights two observations.
(1) Stronger agent configurations reduce interaction, but not monetary cost.
Compared with GPT-6 Luna, GPT-6 Sol and Sonnet 5.5 reduce tool calls by
48.5\% and 51.1\%, and fresh tokens by 9.2\% and 55.1\%, respectively.
Under a common fresh-token mix, list prices imply estimated costs of
18.2$\times$ and 9.0$\times$ Luna's, excluding cache reads
(App.~\ref{app:pricing}).

(2) The workbench lets a lightweight agent match the stronger configurations
in completion. Luna and Sol complete all 12 attempts; Sonnet completes 11.
On these tasks, stronger configurations mainly reduce interaction,
while Luna achieves comparable completion at lower estimated cost.

\subsection{Further Analysis}
\label{sec:further}

\noindent\textbf{Coverage.}
\method contains 44 of the 50 most cited methods, compared with 29
across the three other codebases combined (Fig.~\ref{fig:further}a).
The catalog combines established baselines with recent methods that have
official code, reducing the integration needed to begin a comparison.
This measure captures baseline availability; forecasting quality is
evaluated through the research tasks.

\noindent\textbf{Compact implementations.}
Broader coverage accompanies compact model implementations
(Fig.~\ref{fig:further}b). The median implementation has 144 lines,
close to \texttt{TSLib}'s 145, while the largest has 505 lines,
compared with 2{,}071 in \texttt{TFB}. Shared units consolidate common
computations, and method specifications handle construction and parameter
requirements outside model logic. These statistics describe model code;
the codebases differ in method coverage and supported functionality.

\noindent\textbf{Maintenance through reuse.}
As the catalog grows, extracting shared units helps control code per method
(Fig.~\ref{fig:further}c): one extraction reduced the average function count
from 11.9 to 8.6. A separate refactor reduced maximum function complexity
from 20 to 9 while preserving bit-identical parameters, outputs, and
gradients on reference cases for all \nmodels methods.
Research sessions also expose reusable fixes. Missing guidance, manual
configuration, repeated log polling, and premature test access led to instruction and runner updates; all 17 initial usability failures
were repaired before integration. These cases illustrate how Recursive Improver turns recurring problems into shared improvements,
although they do not isolate the effect of individual checks.

\noindent\textbf{Integration beyond model code.}
Model implementation accounts for only 9\% of method-intake tokens
(Fig.~\ref{fig:further}d). Source reading, catalog search, and verification
account for 28\%, 27\%, and 28\%, respectively. Efficient integration
therefore requires reusing knowledge and verification evidence as well as
code. Knowledge cards retain these findings, helping later agents locate
suitable methods and units without repeating the same investigation.
The mean intake uses 6.1M tokens including cache reads; research-task
costs in Tab.~\ref{tab:main} count fresh tokens and exclude cache reads.

\section{Conclusion}

We presented \method, a shared TSF research workbench that leaves the
agent's model and harness unchanged. It combines domain guidance,
structured knowledge, reusable units, execution, and verified
resource updates. Across the evaluated tasks, it improves task completion and
token efficiency, and supports method development. %
Ablations show complementary benefits from guidance, resource organization,
and shared execution. Retaining source knowledge, implementations, and
experiment records %
allows new agents to reuse verified results and repairs.
\method\ supports cumulative TSF research across agents and
teams, allowing verified improvements to remain available as research
tasks, underlying models, and agent harnesses %
evolve over time.%

\section*{Limitations}

\method currently focuses on time series forecasting. Its guidance,
resource descriptions, and experiment interfaces are designed for TSF
methods and evaluation protocols, so their effectiveness in other research
domains remains untested. Extending the workbench would require adapting
these resources and checks to the tasks and conventions of each domain.
Within the limited search time, method development has not yet produced
a new method that substantially outperforms all leading forecasting
models. The reported improvements therefore demonstrate support for the
evaluated development tasks rather than a broadly superior forecasting
method. More effective search strategies and experimentation over longer
research cycles remain directions for future work.

\section*{Ethical Considerations}

\method records the sources, upstream licenses, and reference revisions
of the methods it incorporates. Datasets retain their original licenses,
and data that cannot be redistributed are obtained from their providers;
all data are public numeric time series without personal information.
Reviews document discrepancies between papers and code together with the
chosen implementation, making these decisions open to inspection.
Experimental results should be interpreted under their recorded protocols;
benchmark rankings do not establish that a method is best across all
settings. Coding agents execute code with network access, so their runs
should remain sandboxed, with external actions requiring explicit
authorization. Agent sessions and model training also consume GPU time
and API tokens; \method reduces these costs by sharing implementations
and lowering token use. \S\ref{sec:main} reports the time and token
cost of each setting, and App.~\ref{app:tasks} gives the hardware and
API prices used for these estimates.

\clearpage  %
\bibliographystyle{acl_natbib}
\bibliography{refs-nolink}

\clearpage
\appendix
\suppressfloats[t]  %
\begingroup
\raggedbottom
\clubpenalty=10000
\widowpenalty=10000
\let\appendixsubsection\subsection
\renewcommand{\subsection}{\Needspace{6\baselineskip}\appendixsubsection}
\renewcommand{\baselinestretch}{1.04}\normalsize
\lstset{basicstyle=\ttfamily\fontsize{10}{11.5}\selectfont,
  columns=fullflexible,keepspaces=true,
  linewidth=\linewidth,xleftmargin=8pt,xrightmargin=8pt,framesep=6pt,
  aboveskip=8pt,belowskip=8pt,breakatwhitespace=true,
  captionpos=t,abovecaptionskip=5pt,belowcaptionskip=7pt}
\section{Framework Structure and Domain Guide}
\label{app:guidance-source}

\subsection{Framework Structure}
The workbench separates agent decisions from shared execution. The agent
reads guidance, selects resources from the catalog, declares an experiment,
and interprets its records. Four modules support this sequence:
\begin{enumerate}[itemsep=4pt,topsep=5pt]
\item \textbf{TSF-Domain Guide:} entry instructions and skills specify workflows, protocol checks, and pitfalls.
\item \textbf{Knowledge Structurer:} cards record the interfaces, configuration rules, and source evidence of each method, dataset, and shared unit in the catalog.
\item \textbf{Experiment Workflower:} presets and method specifications connect each configuration to the shared loading, training, evaluation, and recovery routines.
\item \textbf{Recursive Improver:} usability, complexity, and consistency checks screen new resources and repairs.
\end{enumerate}
The resulting flow is guidance $\rightarrow$ resource selection
$\rightarrow$ configuration $\rightarrow$ execution $\rightarrow$ review
$\rightarrow$ shared updates. The repository keeps models in
\texttt{src/tsflab/models/<model>/}, units in \texttt{models/\_components/}, and workflows in \texttt{.agents/skills/}.
Cards describe resources; specifications, configurations, and tests define
runtime behavior.

\subsection{Entry Instructions and Skills}
Every session starts from one harness-neutral entry file, \texttt{AGENTS.md}
(Claude Code reads the same file through \texttt{CLAUDE.md}). It states the
workbench conventions, module interactions, resource access, and
verification requirements. Task-specific skills provide the steps, checks,
and common pitfalls of each workflow and are loaded as needed. For new-data
benchmarking, these procedures guide the agent through identifying data
fields and evaluation settings, adding the dataset, and inspecting its
splits and scaling before configuring experiments. Listing~\ref{lst:agents} summarizes the entry instructions used
in the benchmarking snapshot.

\begin{lstlisting}[caption={Condensed entry instructions from \texttt{AGENTS.md}.},label={lst:agents}]
# TSFLab Agent Guide (condensed)
Workflows: .agents/skills/; tasks: .agents/tasks/;
contracts: .agents/STANDARDS.md.
## Resource contracts
- Keep models in src/tsflab/models/<model>/.
- Put paper-neutral units in models/_components/.
- Each resource has card.toml, README.md, and optional reference.md.
- Verify source claims; pin revisions and record differences.
- Record executed model admission with `tsf model verify`.
## Progressive access and execution
- Start with `tsf catalog` and `tsf catalog search`.
- Read `tsf catalog show <name>`; deepen with --depth 2|3.
- Use `tsf catalog match <dataset>` for resource selection.
- Inspect splits and scaling with `tsf data splits <dataset>`.
- Register new data with `tsf data add`; execute with `tsf run`.
- The agent owns decisions; APIs own execution guarantees.
## Setup and verification
uv sync --frozen --python 3.12
uv run tsf repo check --audit
uv run tsf model verify --changed
\end{lstlisting}

\begin{sloppypar}
The 17 skills by module: data (\texttt{add-dataset},
\texttt{inspect-dataset}), models (\texttt{add-model},
\texttt{integrate-foundation-model}, \texttt{curate-components},
\texttt{discover-papers}), experiments (\texttt{setup-environment},
\texttt{run-experiment}, \texttt{diagnose-experiment},
\texttt{reproduce-paper-results}, \texttt{analyze-results}), release
(\texttt{submit-results}, \texttt{forecast-realtime-round},
\texttt{publish-weights}), autoresearch (\texttt{run-autoresearch}), and
maintenance (\texttt{audit}, \texttt{handle-contribution}).
\end{sloppypar}

\section{Knowledge Cards}
\label{app:cards}

Cards support summary, card, and source access. These examples retain the
TOML and Markdown formats of the benchmarking snapshot, shortened where marked.
Structured fields, usage guidance, and source evidence are stored separately;
automatic checks and read-only review validate fields and descriptions.

\subsection{Method Card: PatchTST}
\begin{lstlisting}[caption={PatchTST: \texttt{card.toml}.},label={lst:card-method}]
kind = "model"
name = "PatchTST"
fits = ["long-lookback", "long-horizon",
        "weak-cross-correlation"]
fidelity = "reference-checked"
[paper]
title = "A Time Series is Worth 64 Words: ..."
venue = "ICLR 2023"
[code]
url = "https://github.com/yuqinie98/PatchTST"
revision = "204c21ef"
license = "Apache-2.0"
[composition]
normalization = "component:revin"
temporal = "local:patch-transformer-encoder"
channel = "local:channel-independent-shared-weights"
head = "local:flatten-linear-head"
[data_params]
enc_in = { from = "channels" }
patch_len = { from = "seq_len", rule = "at most seq_len" }
[admission]
status = "passed"   # tsf model verify, 2026-10-03
\end{lstlisting}

\Needspace{5\baselineskip}
\begin{lstlisting}[title={\texttt{README.md} (shortened)}]
description: Channel-independent Transformer over overlapping patches of each series,
  with a flatten head and RevIN. Use for long-horizon multivariate forecasting with
  long lookbacks; not for tasks needing cross-channel interaction, exogenous inputs,
  or probabilistic output.
## When to use
- Long horizons where a long lookback helps: patching cuts attention cost.
- Not when cross-channel interaction carries the signal (no channel mixing) or
  calendar marks matter (ignored).
## Configure
- enc_in must equal the data's channel count.
- patch_len must not exceed seq_len.
## Differences
- Clean-room implementation; official code for reference only.
- No self-supervised pretraining or residual attention.
\end{lstlisting}

\subsection{Shared-Unit Card: RevIN}
\begin{lstlisting}[caption={revin: \texttt{card.toml}.},label={lst:card-unit}]
kind = "component"
name = "revin"
slot = "normalization"
role = "Removes each window's level and scale before the
        model and restores them on the forecast."
fits = ["non-stationary", "train-val-level-shift",
        "train-val-scale-shift"]
input = "[batch, time, channels]"
output = "same shape as input"
origin = "RevIN, Kim et al., ICLR 2022"
origin_models = ["patchtst"]
\end{lstlisting}

\Needspace{5\baselineskip}
\begin{lstlisting}[title={\texttt{README.md} (shortened)}]
## What it does
norm:   mu = mean_t(x), sigma = sqrt(var_t(x) + eps),
        y = (x - mu) / sigma, then y * gamma + beta if affine.
denorm: inverts the affine, then z * sigma + mu with the
        statistics cached by the latest norm call (detached).
## When to use
Level or scale drifts across windows or between training and evaluation. Not when
output channels differ from input, or when statistics must carry gradients.
## Interface
RevIN(num_features, eps=1e-5, affine=True,
      subtract_last=False, enabled=True)
forward(values, mode) with mode in {"norm", "denorm"}
\end{lstlisting}

\subsection{Dataset Card: ETTh1}
\begin{lstlisting}[caption={etth1: \texttt{card.toml}.},label={lst:card-data}]
kind = "dataset"
name = "etth1"
domain = "energy"
characteristics = ["non-stationary", "within-train-drift",
                   "regime-changes"]   # tsf data analyze
[source]
url = "https://github.com/zhouhaoyi/ETDataset"
license = "CC-BY-ND-4.0"
[shape]
frequency = "hourly (1h)"
length = 14400   # first 14,400 rows of the CSV
channels = 7
target = "OT"
[protocol]
split = "6:2:2"   # scaling fitted on training rows only
seq_lens = [96, 336, 512]
pred_lens = [96, 192, 336, 720]
\end{lstlisting}

\Needspace{5\baselineskip}
\begin{lstlisting}[title={\texttt{README.md} (shortened)}]
## Protocol and pitfalls
- Truncation: only the first 14,400 rows (20 months) are used; 6:2:2 reproduces the
  12/4/4-month split (7:1:2 results are not comparable).
- drop_last: test keeps its last partial batch, so metrics cover every window; code
  that drops it reports different numbers.
- Distribution shift: OT mean/std is 17.13/9.18 (train) vs. 4.85/3.15 (test).
\end{lstlisting}

\section{Experiment Workflower}
\label{app:runner}

\subsection{Declarative Configurations}
An experiment is a TOML file that composes presets: a base configuration
(training, optimizer, checkpointing, metrics), a dataset preset (path, target,
scaling, split ratio), and a method preset (architecture parameters), followed
by overrides and a sweep. For methods already integrated into the workbench, agents declare the
experiment without writing a separate training script. The runner resolves
the configuration into one run per sweep cell (Listing~\ref{lst:runconfig}). Parameters that a
method card declares as following the data (\texttt{[data\_params]}, \eg
\texttt{enc\_in = \{ from = "channels" \}}) are set from the dataset at run
time, reducing manual changes when reusing a method preset on new data. Each
method specification defines parameter requirements, construction, and
input and output interfaces, connecting the model to the shared loader,
trainer, and evaluator.

\begin{lstlisting}[caption={A run configuration: \texttt{PatchTST} on ETTh1 at four horizons and two seeds.},label={lst:runconfig}]
extends = ["../base.toml",  # training protocol
           "../datasets/etth1.toml",  # split 6:2:2, train-only scaling
           "../models/PatchTST.toml"]  # architecture preset

[task]
seq_len = 336  # override

[sweep]
experiment.random_seed = [0, 1]

[sweep.task]
pred_len = [96, 192, 336, 720]  # 8 runs in total
\end{lstlisting}

\subsection{Command Reference}
Commands are grouped by workflow below. Angle brackets denote
placeholders; bracketed options are optional. The examples document the
interfaces used by the workbench, rather than a single shell script.

\begin{lstlisting}[caption={Resource discovery and dataset commands.},label={lst:commands-catalog}]
# Browse resources, then retrieve only the required detail.
tsf catalog
tsf catalog search <query>
tsf catalog show <name> [--depth 2|3]
tsf catalog match <dataset>
# Register data and inspect its characteristics and protocol.
tsf data add
tsf data analyze
tsf data splits <dataset>
\end{lstlisting}

\begin{lstlisting}[caption={Experiment execution and recovery commands.},label={lst:commands-run}]
# Resolve the run matrix and validate before training.
tsf run <config> --dry-run
tsf run <config>

# Inspect or wait for execution without repeated log polling.
tsf run status <dir>
tsf run wait <dir>

# Stop or recover a run or sweep.
tsf run cancel <dir>
tsf run resume <dir>
\end{lstlisting}

\begin{lstlisting}[caption={Research, result comparison, and verification commands.},label={lst:commands-review}]
# Start a research round and compare results within a protocol.
tsf research start
tsf result report

# Execute model admission and repository audit checks.
tsf model verify --changed
tsf repo check --audit
\end{lstlisting}

\subsection{Budgets and Research Rounds}
An optional execution policy sets budgets (runs, parallel jobs, wall time,
GPU-hours) and resources (GPUs per run, free memory, threads per run); the
scheduler leases GPUs exclusively and counts all attempts against the budget.
Runs attached to a research round (\texttt{tsf research start}) score the
validation split unless overridden, for validation-based search. The method-development protocol is specified in
App.~\ref{app:tasks}.

\subsection{Records and Comparable Results}
Every run writes a record with its resolved configuration, metrics,
checkpoints, and timing, and appends a row to a results table that carries
protocol and model-variant fingerprints (data content, task, evaluation, and
training settings). \texttt{tsf result report} ranks methods only within one
protocol and reports seed coverage and spread. The same records feed the
public \method leaderboard, which ranks methods on fixed benchmark protocols
and on real-time tracks that are forecast before the targets are observed
(\boardurl); the two benchmarking datasets of App.~\ref{app:tasks} are static exports
from these tracks and are evaluated with chronological splits.

\Needspace{9\baselineskip}
\section{Experimental Setup and Evaluation}
\label{app:tasks}

\subsection{Shared Setup}
Codebases are pinned (\texttt{TSLib} at 4e938a1, \texttt{TFB} at a0087ed), and
every setting receives the same prompt, task file, and raw data. A session runs
the agent headless with no additional user instructions beyond the task
prompt, no persistent memory, plug-ins, or MCP servers,
on dedicated NVIDIA RTX 6000 Ada GPUs with network access. Each task uses a
fixed \method snapshot, taken in the order reproduction, benchmarking, and
method development; the released code (\method 1.0.0) is the latest version.

\subsection{Datasets}
\noindent\textbf{Reproduction and development.}
ETTh1~\citep{zhou2021informer} is the hourly energy dataset used in both tasks. We use its first
14,400 rows and seven channels, with 8,640/2,880/2,880 training, validation,
and test rows. Traffic~\citep{wu2021autoformer,chen2001pems} provides the second reproduction dataset; both
reproduction datasets are evaluated at horizons 96, 192, 336, and 720.
Weather~\citep{wu2021autoformer} supplies the second development task, with 21 channels and a
chronological 7:1:2 split. Development uses input 336 and horizon 192 on
both datasets. Model-specific reproduction input lengths and training requirements follow.

\noindent\textbf{New-data benchmarking.}
ERCOT\footnote{\url{https://www.ercot.com}} contains hourly load for eight weather zones, covering 2024-10-01
through 2026-09-30 (17,520 rows). Openmeteo\footnote{\url{https://open-meteo.com}} contains hourly 2\,m temperature
for 82 cities, covering 2025-10-01 through 2026-09-30 (8,760 rows).
Both are fixed CSV exports from the workbench's real-time tracks.
These experiments use historical chronological splits, rather than the
live forecasting protocol of the original tracks. All channels are forecast
with input 168 and horizon 24; scaling is fitted on the training partition
only, and metrics use the normalized scale.

\subsection{Paper Reproduction}
Papers are arXiv forecasting papers first submitted on or after
2026-06-01. We kept papers that have no official code, are absent from every
compared codebase, report per-horizon MSE and MAE on ETTh1 and Traffic at
horizons 96, 192, 336, and 720 under the standard protocol, give enough training detail, and train all four horizons in under one
GPU-hour. Each session has four hours, one GPU, and a ready environment.
\begin{itemize}[leftmargin=*,itemsep=1pt,topsep=2pt]
\item \textbf{RhyMix} \citep{shevtekar2026rhymix}: input 96; cyclic and multi-scale convolution paths with gated forecasting heads; Huber loss. Each cell uses five seeds to match the paper's five-run means. Batch size and gate sizes are not specified.
\item \textbf{StateFlow} \citep{gharwi2026stateflow}: input 96; recurrent hidden and residual-memory states with a chunk decoder. Part of the cell is defined in an earlier paper; optimizer and two-stage schedule are unspecified.
\item \textbf{AdaRDiff} \citep{laglil2026adardiff}: input 720; a learnable reversible residual transform around a linear backbone. Training runs 30 residual-loss then 10 forecast-loss epochs. The paper gives only the search grid.
\item \textbf{CoRe} \citep{lin2026core}: input 96; coherence and low-rank channel-alignment losses applied to \texttt{iTransformer}. Loss weight, rank, and learning rate are unspecified. It is related to the existing FreDF method, so the task tests whether agents reuse it.
\end{itemize}

\subsection{Benchmarking on New Data}
The task uses the ERCOT and Openmeteo exports described above. The eight
methods are three established baselines contained in all three codebases and
the five most cited forecasting methods of 2026 venues that \method contains and
\texttt{TSLib} and \texttt{TFB} do not (Semantic Scholar, 2026-10-09):
\begin{itemize}[itemsep=3pt,topsep=4pt]
\item \texttt{DLinear} decomposes the input into trend and seasonal parts and forecasts each with a linear layer. \texttt{PatchTST} applies a channel-independent Transformer to patches of each series; \texttt{iTransformer} instead treats each variate's input history as one token and attends across variates.
\item \textbf{TimeAlign} trains a patch MLP with a future-reconstruction branch and aligns history and future representations at local and global levels; inference uses the patch MLP alone.
\item \textbf{GTR} retrieves segments from a learned full-cycle memory and fuses them with the input by 2D convolution, followed by a residual MLP.
\item \textbf{TimeMosaic} adapts patch size to the information density of each input region and uses segment-specific prompts and linear heads for forecasting.
\item \textbf{PHAT} buckets each variable by FFT-derived phase and cycle, with positive--negative attention per bucket.
\item \textbf{EMAformer} augments variate tokens with channel, cycle-phase, and joint channel--phase embeddings.
\end{itemize}
References for these methods are listed in \S\ref{sec:tasks}.

This task therefore measures the combined benefit of available method
implementations and the experiment workflow; method availability differs
across codebases. The fixed protocol is:
\begin{itemize}[leftmargin=*,itemsep=3pt,topsep=5pt]
\item \textbf{Data.} Chronological 70/10/20 split; train-only scaling; all channels; input 168 and horizon 24.
\item \textbf{Training.} Paper or official-code architecture settings; at most ten epochs; validation early stopping with patience 3; batch size 32; seed 42.
\item \textbf{Evaluation.} Every test window (stride 1); MSE and MAE on the normalized scale.
\item \textbf{Budget.} Six hours per session, including environment installation.
\end{itemize}

\subsection{Method Development Pilot}
On ETTh1 and Weather, the agent searches for a method on validation data under a fixed protocol:
\begin{itemize}[leftmargin=*,itemsep=3pt,topsep=5pt]
\item \textbf{Data.} All channels; input 336; horizon 192; train-only standardization; seed 2021.
\item \textbf{Training.} MSE loss; Adam, lr $10^{-3}$; at most ten epochs; patience 3; best-validation checkpoint.
\item \textbf{Selection.} Validation-only candidate selection; the frozen method's saved checkpoint is tested once.
\item \textbf{Budget.} 90 minutes; two GPUs; at most 12 fits and four search iterations per session.
\end{itemize}

\subsection{Evaluation Metrics}
\label{app:measure}

\noindent\textbf{Completion.}
For reproduction and the ablations, Completed counts accepted sessions out of
$N$ ($x/N$). A script first checks that every required output exists. Two
read-only reviewer agents then answer a fixed checklist: where the split is
made, where scaling statistics are fitted, which data select checkpoints,
whether every test window is evaluated, and whether reported values come from
executed runs; a third reviewer decides disagreements. A session is accepted if
it covers the complete requested matrix with executed metrics, the prescribed
splits, train-only scaling, the common evaluation over all windows, and no
test-informed configuration selection; agreement with the paper's MSE is evaluated separately and is not an acceptance
threshold. For
benchmarking, Completed counts accepted method--dataset results out of 48 (eight methods,
two datasets, three attempts). A result is accepted if its MSE and MAE are
finite, the reviews (after any tie-break) support the protocol, the code implements the method's
defining mechanisms, and the metrics have execution evidence in the run records.

\noindent\textbf{Accuracy.}
For $W$ test windows, horizon $H$, and $C$ channels on the normalized scale,
\[
\mathrm{MSE}=\frac{1}{WHC}\sum_{w,h,c}\big(\hat{y}_{w,h,c}-y_{w,h,c}\big)^2,\quad
\mathrm{MAE}=\frac{1}{WHC}\sum_{w,h,c}\big|\hat{y}_{w,h,c}-y_{w,h,c}\big|.
\]
The relative error of a reproduced result is
$\epsilon = 100\,|\mathrm{MSE}_{\mathrm{run}}-\mathrm{MSE}_{\mathrm{paper}}|/
\mathrm{MSE}_{\mathrm{paper}}$.

\noindent\textbf{Cost.}
For a setting with attempts $s\in S$ and $n$ completed outcomes,
\[
\mathrm{Time}=\frac{1}{n}\sum_{s\in S}T_s,\quad
\mathrm{Tokens}=\frac{1}{n}\sum_{s\in S}K_s,\quad
\mathrm{Calls}=\frac{1}{n}\sum_{s\in S}A_s,
\]
where $T_s$ is wall time, $A_s$ the number of agent tool calls, and $K_s$ the
fresh tokens of attempt $s$: non-cached input, cache-write input, and output,
excluding cache reads. $S$ includes failed attempts. For reproduction,
$n$ counts accepted sessions; for benchmarking, it counts accepted
method--dataset results. These denominators differ, so costs are compared
within each task. The method-intake analysis separately reports tokens
including cache reads, as stated in the main text.

\noindent\textbf{Code changes.}
Let $F_0$ and $F_1$ be the functions of the initial and final source of a
session and $\mathrm{CC}(f)$ the cyclomatic complexity of $f$ (Radon 6.0.1),
with $\mathrm{CC}_0(f)=0$ for $f\notin F_0$. Cyclomatic complexity counts
independent execution paths through a function; conditionals and loops
typically increase it. Then
\[
\Delta\mathrm{LOC}=L^{+}-L^{-},\quad
\Delta\mathrm{Funcs}=|F_1\setminus F_0|,\quad
\Delta\mathrm{CC}=\sum_{f\in F_1}\max\big(0,\mathrm{CC}_1(f)-\mathrm{CC}_0(f)\big),
\]
where $L^{+}$ and $L^{-}$ are added and removed Python source lines. The source
includes experiment scripts and excludes official repositories cloned into the
workspace, installed environments, and generated run records. Code metrics are
averaged over sessions within each task and then across tasks with equal weight.

\subsection{API Price Assumptions}
\label{app:pricing}
The cost comparison in \S\ref{sec:agents} uses API list prices
checked on October 13, 2026 (USD per million tokens). GPT-6 rates are short-context. Sources: the official
\href{https://developers.openai.com/api/docs/models/gpt-6-luna}{GPT-6 Luna},
\href{https://developers.openai.com/api/docs/models/gpt-6-sol}{GPT-6 Sol}, and
\href{https://platform.claude.com/docs/en/models/sonnet-5-5/overview}{Sonnet 5.5}
model pages.

\begin{center}
\begin{tabular}{lrrrr}
\toprule
Model & Input & Output & Cache write & Cache read \\
\midrule
GPT-6 Luna & 0.10 & 0.50 & 0.125 & 0.01 \\
GPT-6 Sol & 2.00 & 10.00 & 2.50 & 0.20 \\
Sonnet 5.5 & 2.00 & 10.00 & 2.50 & 0.10 \\
\bottomrule
\end{tabular}
\end{center}
Sonnet's cache-write rate above is for five-minute caching; one-hour
writes cost \$4.00 per million tokens. For non-cached input $I$,
cache-write input $W$, and output $O$, the estimated fresh-token charge is
\[
\widehat{C}_{\mathrm{fresh}}=
\frac{p_I I+p_W W+p_O O}{10^6}.
\]
Cache reads are excluded from fresh tokens and this estimate; adding
$p_R R/10^6$ gives the token charge including $R$ cache-read tokens.
The comparison assumes the same fresh-token composition
across agents, using five-minute cache writes for Sonnet. Each fresh-token rate for Sol and
Sonnet is then 20 times Luna's. Using Tab.~\ref{tab:agents}, the estimated
ratios are $20(0.394/0.434)\approx18.2$ and $20(0.195/0.434)\approx9.0$, respectively.
These are normalized estimates, not invoices; actual
charges depend on token composition, context tier, and caching.

\endgroup

\end{document}